%% file: main.tex
\pdfoutput=1
\documentclass[11pt]{article}

\usepackage[final]{acl}

\input{preamble}

\usepackage{times}
\usepackage{latexsym}

\usepackage[T1]{fontenc}
\usepackage[utf8]{inputenc}

\usepackage{microtype}

\usepackage{inconsolata}

\title{Multimodal Instruction Tuning with Conditional Mixture of LoRA}

\author{Ying Shen$^{\spadesuit}$ \quad Zhiyang Xu$^{\spadesuit}$  \\ 
    \textbf{Qifan Wang}$^{\diamondsuit}$ \quad \textbf{Yu Cheng}$^{\clubsuit}$ \quad \textbf{Wenpeng Yin}$^{\heartsuit}$ \quad \textbf{Lifu Huang}$^{\spadesuit}$\\
  $^{\spadesuit}$Virginia Tech \quad $^{\diamondsuit}$Meta AI \quad $^{\clubsuit}$The Chinese University of Hong Kong \\ $^{\heartsuit}$The Pennsylvania State University \\
  $^{\spadesuit}$\texttt{\{yings, zhiyangx, lifuh\}@vt.edu} \quad $^{\diamondsuit}$\texttt{wqfcr@meta.com} \\ $^{\clubsuit}$\texttt{chengyu@cse.cuhk.edu.hk} \quad $^{\heartsuit}$\texttt{wfy5054@psu.edu}
  \\ 
  }

\begin{document}
\maketitle

\input{sec/0_abstract}
\input{sec/1_intro}
\input{sec/2_related}

\input{sec/3_blackground}

\input{sec/4_method}

\input{sec/5_exp}

\input{sec/6_results}

\input{sec/7_conclusion}
\input{sec/8_limitations}
\input{sec/9_acknowledgements}

\bibliography{anthology,custom}
\appendix
\input{sec/X_suppl}

\end{document}

%% file: preamble.tex
\usepackage{colortbl}
\usepackage{multirow}
\usepackage{xcolor}
\usepackage{graphicx}
\usepackage{nicematrix}
\usepackage{subcaption}
\usepackage{array}
\usepackage{booktabs}
\usepackage{amsmath}
\usepackage{environ}
\usepackage{nicematrix,tikz}

\usepackage{amssymb}% http://ctan.org/pkg/amssymb
\usepackage{pifont}% http://ctan.org/pkg/pifont

\newcommand*\modelname{Conditional Mixture-of-LoRA}
\newcommand*\shortmodelname{MixLoRA}

\newcolumntype{g}{>{\columncolor[gray]{0.8}}c}

\newcommand {\first}[1]{{\textbf{#1}}}

\definecolor{loraB}{HTML}{548235}
\definecolor{loraA}{HTML}{0070C0}

\NewDocumentCommand{\ying}{ mO{} }{\textcolor{teal}{\textsuperscript{\textit{Ying}}\textsf{\textbf{\small[#1]}}}}

\NewDocumentCommand{\lifu}{ mO{} }{\textcolor{blue}{\textsuperscript{\textit{Lifu}}\textsf{\textbf{\small[#1]}}}}

\NewEnviron{gather+}[1][1]{%
  \begin{equation}
  \scalebox{#1}{$\begin{gathered}\BODY\end{gathered}$}
  \end{equation}
}

%% file: sec/0_abstract.tex
\begin{abstract}

Multimodal Large Language Models (MLLMs) have demonstrated remarkable proficiency in diverse tasks across different domains, with an increasing focus on improving their zero-shot generalization capabilities for unseen multimodal tasks. 
Multimodal instruction tuning has emerged as a successful strategy for achieving zero-shot generalization by fine-tuning pre-trained models on diverse multimodal tasks through instructions.
As MLLMs grow in complexity and size, the need for parameter-efficient fine-tuning methods like Low-Rank Adaption (LoRA), which fine-tunes with a minimal set of parameters, becomes essential.
However, applying LoRA in multimodal instruction tuning presents the challenge of task interference, which leads to performance degradation, especially when dealing with a broad array of multimodal tasks.
To address this, this paper introduces a novel approach that integrates multimodal instruction tuning with \modelname{} (\shortmodelname{}). It innovates upon LoRA by dynamically constructing low-rank adaptation matrices tailored to the unique demands of each input instance, aiming to mitigate task interference.
Experimental results on various multimodal evaluation datasets indicate that \shortmodelname{} not only outperforms the conventional LoRA with the same or even higher ranks, demonstrating its efficacy and adaptability in diverse multimodal tasks\footnote{Our code is publicly available at \url{https://github.com/VT-NLP/MixLoRA}}.

\end{abstract}

%% file: sec/1_intro.tex
\section{Introduction}

The advent of Multimodal Large Language Models (MLLMs) \cite{li2023blip,liu2023llava,pmlr-v202-driess23a,instructblip} have revolutionized the field of artificial intelligence, demonstrating remarkable capabilities in processing and integrating information from various modalities, notably text and image.
A key focus in advancing MLLMs is to enhance zero-shot generalization to novel multimodal tasks. In this pursuit, multimodal instruction tuning, which fine-tunes pre-trained models with diverse, instruction-based multimodal tasks, has demonstrated its efficacy in facilitating zero-shot generalization to unseen multimodal problems \cite{xu-etal-2023-multiinstruct,liu2023llava,ye2023mplug}.

\input{figures/overview_fig}

Concurrently, the growing complexity and scale of MLLMs have spurred the development of various parameter-efficient fine-tuning (PEFT) techniques \cite{lee2019would,hu2021lora,li2021prefix,karimi2021compacter,guo2021parameter,zaken2022bitfit}. Among these, Low-Rank Adaption (LoRA) \cite{hu2021lora} has emerged as a powerful PEFT method that fine-tunes large pre-trained models by updating a small amount of injected adaption parameters. 
However, in multimodal instruction tuning, the effectiveness of conventional PEFT methods like LoRA diminishes due to their reliance on adjusting a limited portion of shared parameters to simultaneously accommodate diverse tasks, leading to task interference -- a problem well-studied in multi-task learning \cite{yu2020gradient,liu2021conflict,navon2022multi}, but insufficiently investigated in the context of parameter-efficient multimodal instruction tuning.
The diverse nature of multimodal tasks significantly increases the risk of task interference. For instance, using the same limited set of adaptation parameters for distinct tasks like OCR and domain-specific classification can cause conflicting updates, potentially leading to suboptimal performance.
Our research seeks to explore and address task interference in parameter-efficient multimodal instruction tuning. Specifically, we aim to answer two critical research questions: (1) Does task interference exist in parameter-efficient multimodal instruction tuning? (2) How can we effectively mitigate this issue for robust and versatile performance across various multimodal tasks?

To answer the first question, we investigate the task-interference issue in parameter-efficient multimodal instruction tuning from the perspective of gradient direction ~\cite{liu2021conflict} in Section~\ref{subsec:task_inteference}. Our observations highlight notable task interference in this context, underscoring the necessity for more effective adaptation strategies to ensure robust and versatile performance across diverse multimodal tasks. 
In response to our second question, this paper proposes a novel multimodal instruction tuning framework -- \modelname{} (\shortmodelname{}), designed to mitigate the task interference issue.
As shown in Figure~\ref{fig:overall}, unlike conventional LoRA which uses shared low-rank adaptation matrices $A$ and $B$ across all tasks and instances, \shortmodelname{} dynamically constructs low-rank adaptation matrices $A$ and $B$ tailored to each input instance, by selecting their decomposition factors from two collections. 
\shortmodelname{} introduces a dynamic factor selection mechanism, incorporating two Independent Factor Selection (IFS) routers and a Conditional Factor Selection (CFS) router. The two IFS routers independently select appropriate factors to dynamically construct LoRA $A$ and $B$ matrices tailored to each input. 
The CFS router further refines the selection for LoRA $B$ based on the factors chosen for LoRA $A$, ensuring that the factors selections for LoRA $A$ and $B$ are not only tailed to input but also cohesively aligned.

To validate the effectiveness of \shortmodelname{}, we conduct extensive experiments on MME~\cite{fu2023mme}, a comprehensive multimodal evaluation benchmark, and seven additional multimodal evaluation datasets that focus on various capabilities. 
Experimental results demonstrate that \shortmodelname{}, with its dynamic factor selection approach, consistently outperforms LoRA across various multimodal tasks when using the same number of ranks and remains competitive or superior even against LoRA with a higher rank number. 
This effectiveness is attributed to the dynamic factor selection mechanism and its ability to generalize to unseen tasks through adaptive factor activation, underscoring the potential of \shortmodelname{} to generalize and perform effectively on unseen multimodal tasks.

Our contributions are summarized as follows:
(1) We empirically investigate and demonstrate the existence of task interference in parameter-efficient multimodal instruction tuning. 
(2) We propose the \modelname{} (\shortmodelname{}) framework, aimed at alleviating task interference by dynamically constructing low-rank adaptation matrices for various inputs.
(3) Comprehensive experiments demonstrate the effectiveness of \shortmodelname{}, outperforming LoRA across various unseen multimodal tasks at equal or even higher ranks.

%% file: figures/overview_fig.tex
\begin{figure*}[!tbh]
    \centering
\includegraphics[width=\linewidth]{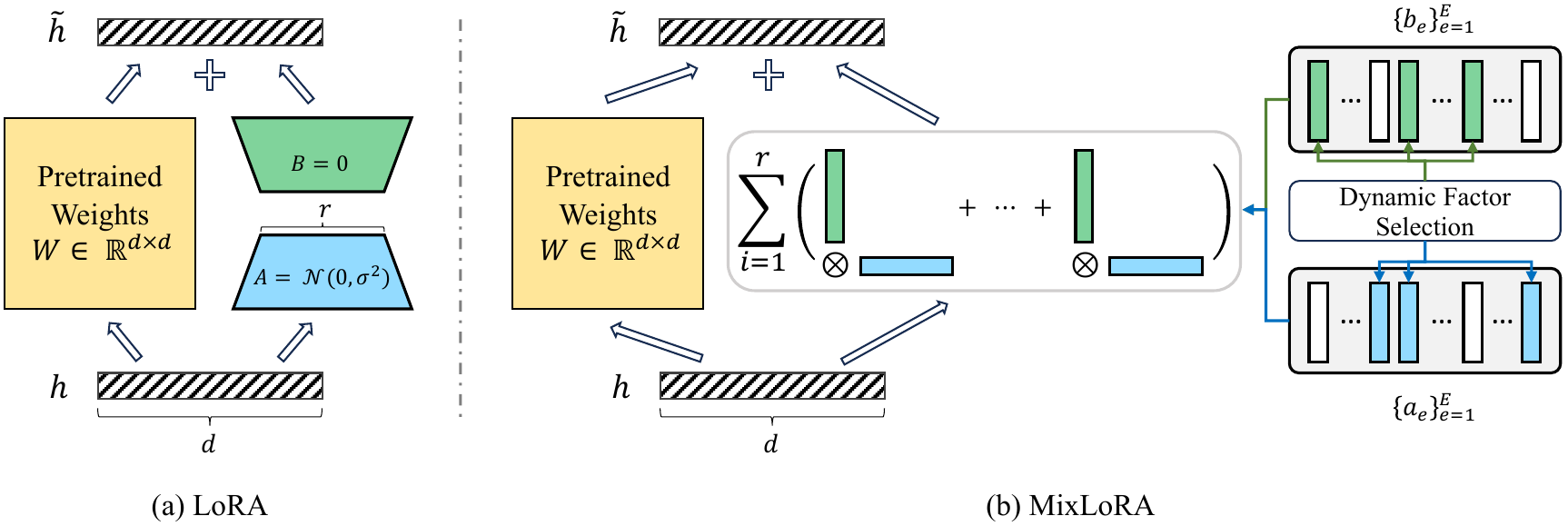}
    \caption{\textbf{Comparative Overview of LoRA and \shortmodelname{}.} \textit{Left}: The conventional LoRA with static low-rank decomposition matrices $BA$. \textit{Right}: \shortmodelname{} treats the low-rank decomposition factors as experts that can be selectively combined through a Dynamic Factor Selection module, enabling the construction of varied low-rank decomposition matrices $A$ and $B$ tailored to varying input scenarios. The selected factors are visually distinguished by color coding: \textcolor{loraB}{green} for $B$ and \textcolor{loraA}{blue} for $A$.
}
    \label{fig:overall}
\end{figure*}

%% file: sec/2_related.tex
\section{Related Work}

\paragraph{Multimodal Instruction tuning}

Instruction tuning \cite{wei2021flan} significantly improves the generalization of large language models to unseen tasks based on natural language instructions~\cite{Ouyang2022instruct-gpt,taori2023alpaca}. 
With the advent of multimodal large language models, the scope of instruction tuning has expanded to encompass multimodal and vision tasks, %\cite{xu-etal-2023-multiinstruct,liu2023llava,ye2023mplug}, 
facilitated by the development of diverse multimodal instruction datasets, including  
both machine-generated~\cite{liu2023llava,zhao2023svit,zhu2023minigpt,yin2023lamm,li2023stablellava,ye2023mplug} and human-annotated~\cite{xu-etal-2023-multiinstruct}. Recently, Vision-Flan~\cite{visionFlan2023} stands out as a comprehensive human-annotated visual instruction tuning dataset, covering a wide range of 187 tasks, making it ideal for our training.

\paragraph{Parameter-efficient fine-tuning (PEFT)}

Parameter-efficient fine-tuning (PEFT) \cite{lee2019would,hu2021lora,li2021prefix,karimi2021compacter,guo2021parameter,zaken2022bitfit} strategies have become key in efficiently adapting large pre-trained models to various downstream tasks with minimal parameter adjustments.
These methods typically involve freezing most of the pre-trained model while fine-tuning a small fraction of the parameters to facilitate the adaptation process. 
Among these, LoRA \cite{hu2021lora} demonstrates competitive trade-offs between performance and parameter efficiency, making it widely adopted.
PEFT methods typically utilize shared adaptation parameters across diverse tasks or train task-specific adapters.
However, when applied to multimodal instruction tuning, which requires simultaneous adaptation to diverse instruction tasks, PEFT can encounter task interference, highlighting the need for more adaptable and versatile PEFT methods to adeptly handle the complexities of multimodal instruction tuning.

\paragraph{Task Interference}

Task interference \cite{crawshaw2020multi} is a notable challenge in multi-task learning, where simultaneous training on multiple tasks can lead to performance decline due to conflicting gradients among tasks \cite{yu2020gradient,liu2021conflict,navon2022multi}.
To mitigate task interference in multi-task learning, researchers have explored various strategies, including dynamic adjustment of task loss contributions 
\cite{chen2018gradnorm,sener2018multi,liu2019end} and parameter partitioning ~\cite{maninis2019attentive,bragman2019stochastic,strezoski2019many,zhang2020share}. Despite the established understanding of task interference in multi-task learning, its presence and implications in instruction tuning, particularly in multimodal contexts, remain under-explored. 
Given the intrinsic complexity and diversity of multimodal instruction-based tasks, substantial task interference is likely to exist in multimodal instruction-tuning scenarios. Our research delves into this area, specifically investigating task interference within parameter-efficient multimodal instruction tuning.

\input{figures/task_interference}

%% file: figures/task_interference.tex
\begin{figure*}[tbh!]
    \centering
    \begin{subfigure}[b]{0.495\linewidth}
        \includegraphics[width=\linewidth]{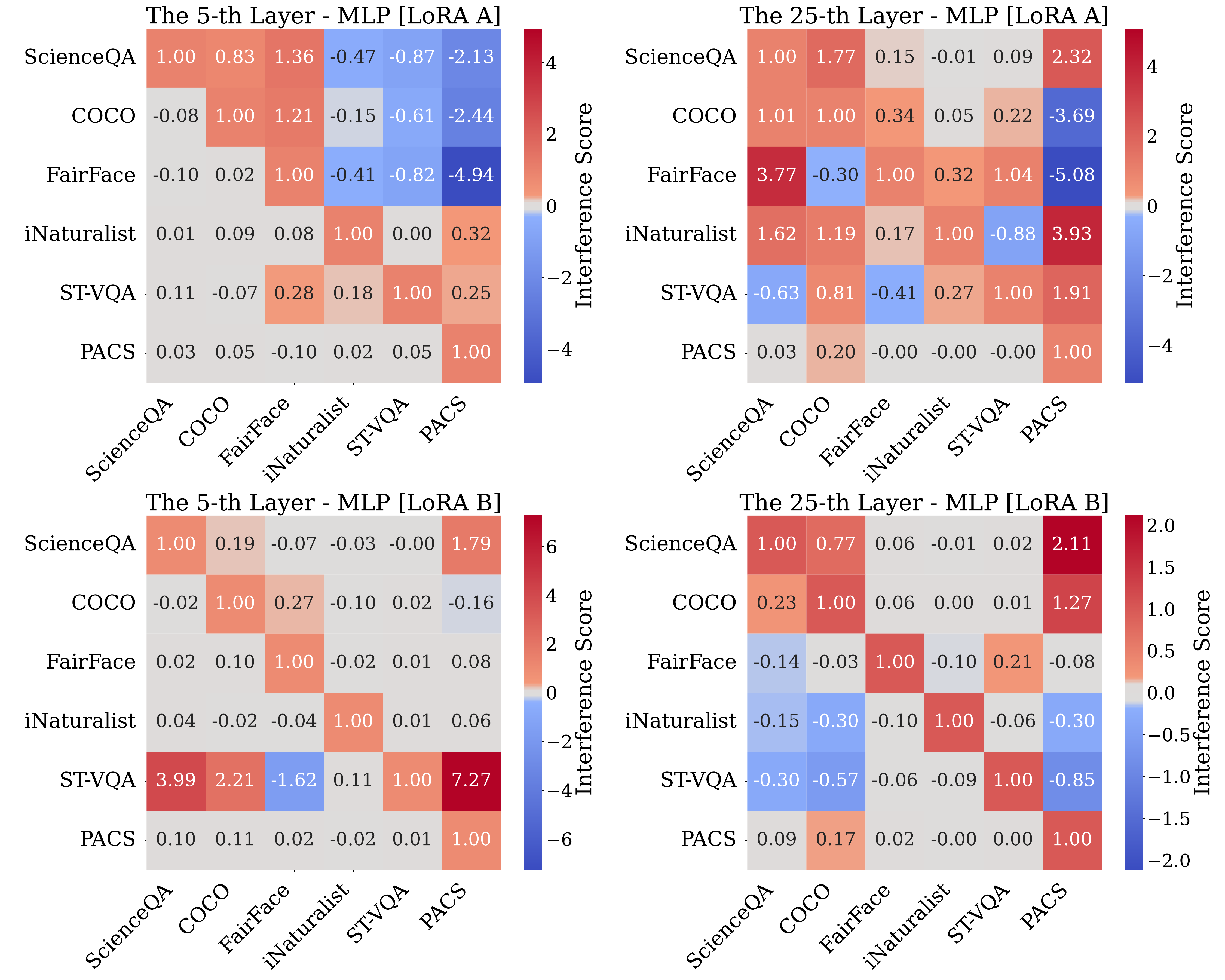}
        \caption{MLP}
        \label{fig:ti_mlp}
    \end{subfigure}
    % \hfill % adds horizontal space between the two subfigures
    \begin{subfigure}[b]{0.495\linewidth}
        \includegraphics[width=\linewidth]{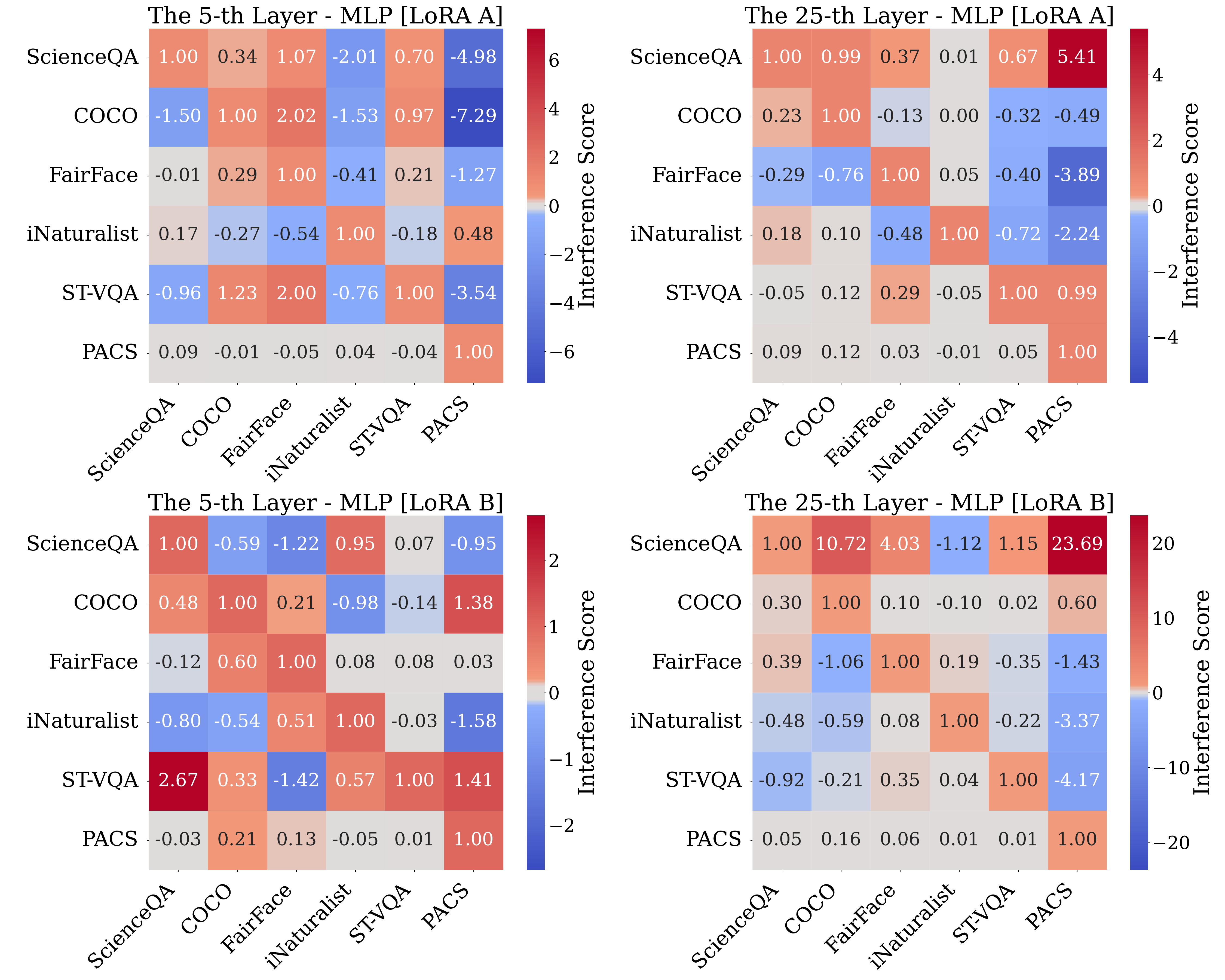}
        \caption{Self-Attention}
        \label{fig:ti_attn}
    \end{subfigure}
    \caption{\textbf{The Task Interference Score $\mathcal{I}$ for LoRA decomposition matrices $A$ and $B$.} Each cell in the heatmap corresponds to the average interference score $\mathcal{I}_{i,j}$ of task $j$ (column) on the task $i$ (row). A blue hue indicates a negative impact of task $j$ on task $i$, whereas a red hue signifies a positive impact.}
    \label{fig:ti}
\end{figure*}

%% file: sec/3_blackground.tex
\section{Task Interference in Multimodal Instruction Tuning with LoRA}

\subsection{Background: Low-Rank Adaptation} Low-Rank Adaptation (LoRA) \cite{hu2021lora} is a parameter-efficient fine-tuning method that fine-tunes only the trainable rank decomposition matrices injected in each layer of the Transformer \cite{vaswani2017attention}.
As illustrated in Figure~\ref{fig:overall} (a), consider a linear layer, represented by $\tilde{h} = W h$, where $W \in \mathbb{R}^{d_{out} \times d_{in}}$ denotes the pre-trained weight, with $d_{in}$ and $d_{out}$ being the input and output dimensions, respectively. LoRA modifies the model parameters by injecting low-rank decomposition matrices as the weight adjustment matrices, which can be expressed as:
% \begin{gather+}[0.8]
\begin{align}
    \tilde{h} = W h + \Delta W h = W h + \alpha \cdot BA h,
    \label{eq:lora}
\end{align}
% \end{gather+}
where $\Delta W = BA$ represents the trainable weight adjustment matrices formed by low-rank matrices $A \in \mathbb{R}^{r \times d_{in}}$ and $B \in \mathbb{R}^{d_{out} \times r}$, with the rank $r \ll \min(d_{in}, d_{out})$. The scalar $\alpha \geq 1$ controls the influence of the weight adjustment matrices. 
During fine-tuning, 
only these low-rank decomposition matrices, referred to as LoRA $A$ and LoRA $B$ throughout this paper, are updated, allowing for rapid, task-specific adaptation by training distinct LoRA $A$ and $B$ for each downstream task.

\subsection{Investigating Task Interference in Multimodal Instruction Tuning}
\label{subsec:task_inteference}

Our study delves into task interference in parameter-efficient multimodal instruction tuning by analyzing gradient direction conflicts between task pairs. For each task pair $i$ and $j$, we first estimate the change in loss $L_i$ of task $i$, when optimizing the shared parameters $\theta$ according to the loss $L_j$ of task $j$, following \cite{zhu2022uni}:
\begin{gather+}[0.84]
    \Delta_j L_i(x_i) = \mathbb{E}_{x_j} \bigg( L_i(x_i; \theta) - L_i(x_i; \theta - \lambda \frac{\nabla_{\theta} L_j(x_j)}{\| \nabla_{\theta} L_j(x_j) \|} ) \bigg)  \\ 
    \approx \lambda \mathbb{E}_{x_j} \left( \frac{\nabla_{\theta} L_j(x_j)}{\| \nabla_{\theta} L_j(x_j) \|}^T \nabla_{\theta} L_i(x_i) \right), \label{eq:L}
\end{gather+}
where $x_i$ and $x_j$ are sampled training batches for tasks $i$ and $j$, and $\lambda$ is the learning rate. 

The interference of task $j$ on task $i$ is then quantified as follows:
% \begin{gather+}[0.8]
\begin{align}
    \mathcal{I}_{i, j} = \mathbb{E}_{x_i} \left(\frac{\Delta_j L_i(x_i)}{\Delta_i L_i(x_i)}\right).
    \label{eq:I}
\end{align}
% \end{gather+}

Here, a positive $\mathcal{I}_{i, j}$ suggests aligned gradient directions between tasks $i$ and $j$, while a negative value implies divergent gradient directions, indicating that task $j$ adversely impacts task $i$.

We conduct experiments on the fine-tuned LLaVa \cite{liu2023llava} model using LoRA with a rank of 4, computing the task interference among six diverse tasks from Vision-Flan~\cite{visionFlan2023}, including 
``ScienceQA''~\cite{lu2022learn} (for ``Complex Reasoning''),``COCO''~\cite{lin2014microsoft} (for ``Coarse-grained Perception''), ``FairFace''~\cite{karkkainenfairface} (for ``Fine-grained Perception''), ``iNaturalist''~\cite{van2018inaturalist} (for ``Knowledge Intensive''), ``ST-VQA''~\cite{biten2019scene} (for ``OCR''), and ``PACS''~\cite{li2017deeper} (for ``Domain specific'').
We compute the average task interference matrix $\mathcal{I}$ based on the gradients concerning LoRA $A$ and $B$, across various layers. 
Figure \ref{fig:ti} shows the task interference score for LoRA $A$ and $B$ at the 5-th and 25-th Transformer Layer for both MLP (Figure \ref{fig:ti_mlp}) and Self-Attention (Figure \ref{fig:ti_attn}). 

Our results reveal notable task interference at both shallow and deep Transformer layers for LoRA $A$ and $B$. For instance, as shown in Figure \ref{fig:ti_attn}, at the $5$-th layer for LoRA $A$, the domain-specific classification task ``PACS'' negatively impacts ``COCO'', a coarse-grained perception task, with a negative interference score of $-7.3$. Meanwhile, positive influences are also observed. For example, Figure \ref{fig:ti_mlp} shows that at the $5$-th layer for LoRA $B$, ``PACS'' positively affects the OCR task "ST-VQA".
The presence of both positive and negative interference suggests complex dynamics among instruction tasks: positive scores (in red),  suggest that the learning of one task can enhance the performance of another, while negative scores (in blue), imply that one task's learning can hinder another.
These findings highlight notable task interference in parameter-efficient multimodal instruction tuning and reinforce the need for effective adaption methods to ensure robust and versatile performance across diverse multimodal tasks.

%% file: sec/4_method.tex
\section{\modelname{}}

Inspired by the concept of Mixture-of-Experts \cite{shazeer2016outrageously}, we propose \modelname{} (\shortmodelname{}) which leverages low-rank decomposition factors as dynamically chosen experts to construct tailored decomposition matrices $A$ and $B$ for specific input instances. 
\shortmodelname{} facilitates dynamic processing pathways for varying input instances, thereby enhancing the efficacy in handling diverse and complex multimodal instruction tasks.

The core of \modelname{} lies in representing the weight adjustment matrices $\Delta W$ from Equation \ref{eq:lora} via tensor decomposition:
% \begin{gather+}[0.8]
\begin{align}
    \Delta W = BA = \sum_{i=1}^r b_i \otimes a_i,
    \label{eq:mixlora}
\end{align}
% \end{gather+}
where $\otimes$ denotes the outer product and $\{a_i, b_i\}_{i=1}^r, a_i \in \mathbb{R}^{d_{in} \times 1}, b_i \in \mathbb{R}^{d_{out} \times 1}$ are the rank $r$ decomposition factors of $\Delta W \in \mathbb{R}^{d_{out} \times d_{in}}$.

Leveraging the concept that $\Delta W$ can be expressed as the sum of outer products of low-rank decomposition factors $a_i$ and $b_i$, \shortmodelname{} introduces a \textbf{Dynamic Factor Selection} module. This module dynamically constructs unique $\Delta W$ for specific inputs by selecting $r$ appropriate factors from an expanded pool of decomposition factors $\{a_e\}_{e=1}^E, \{b_e\}_{e=1}^E, E > r$, as shown in Fig. \ref{fig:overall} (b). 
Here, the number of factors $E$ is much larger than the rank $r$. Following LoRA~\cite{hu2021lora}, we use a random Gaussian initialization for $\{a_e\}_{e=1}^E$ and zero for $\{b_e\}_{e=1}^E$.

\subsection{Dynamic Factor Selection}

The Dynamic Factor Selection module uses two main components to dynamically constructs LoRA $A$ and $B$. First, two \textbf{Independent Factor Selection (IFS)} routers (Section~\ref{sec:ifs}), independently select $r$ relevant factors to form adaptation matrices LoRA $A$ and $B$, ensuring precise, instance-specific adaptations.
Second, a \textbf{Conditional Factor Selection (CFS)} router (Section~\ref{sec:cfs}) further refines the selection for LoRA $B$ by conditioning the selection for $B$ also on the factors chosen for LoRA $A$, promoting a coherent adaptation process.

\subsubsection{Independent Factor Selection}
\label{sec:ifs}

\input{figures/routing}
\shortmodelname{} employs two Independent Factor Selection (IFS) routers, $R^A_{\text{IFS}}(\cdot)$ and $R^B_{\text{IFS}}(\cdot)$, to select $r$ relevant factors for LoRA $A$ and $B$, respectively, as shown in Figure \ref{fig:routing}.
IFS routers employ an instance-based routing method, which is more memory-efficient than conventional input-token-based routing, for selecting $r$ decomposition factors. The routing strategy can be expressed as:
\begin{align}
    R^A_{\text{IFS}}(h) = \text{Avg}(h),
\end{align}
% \end{gather+}
where $\text{Avg}(\cdot)$ averages across the sequence dimension of the hidden states $h \in \mathbb{R}^{seq \times d_{in}}$ from the preceding layer, resulting $ R^A_{\text{IFS}}(h) 
 \in \mathbb{R}^{d_{in}}$.

\paragraph{Factor Selection Process}
The factor selection process involves calculating vectors $g_A \in \mathbb{R}^E$ and $g_B \in \mathbb{R}^E$ to selectively identify specific subsets of decomposition factors from the set $\{a_e\}_{e=1}^E$ and $\{b_e\}_{e=1}^E$, respectively. 
To compute $g_A$, input $R^A_{\text{IFS}}(h) \in \mathbb{R}^{d_{in}}$ is processed through a dense layer with weights $W_A \in \mathbb{R}^{E \times d_{in}}$, followed by a softmax normalization and top-$k$ selection:
% \begin{gather+}[0.8]
\begin{align}
    g_A = \text{top}_r(\text{softmax}(W_A \cdot R^A_{\text{IFS}}(h))).
\end{align}
% \end{gather+}

This procedure ensures the selection of $r$ factors for LoRA $A$, with $g_A[i]=1$ indicating the selection of factor $i$. The same process is applied to determine $g_B$ for LoRA $B$.

\subsubsection{Conditional Factor Selection}
\label{sec:cfs}

While the factors for LoRA $A$ and $B$ have been independently selected so far, we hypothesize that an interdependence exists between the selections for LoRA $A$ and $B$, which can be harnessed to improve the model's overall adaptability and performance. 
To leverage this relationship, we propose a Conditional Factor Selection (CFS) strategy, wherein the selection of factors for the projection-up weight of LoRA $B$ is also influenced by the factors chosen for the projection-down weight of LoRA $A$.

With the IFS router, LoRA $A$ is assembled from chosen decomposition factors, denoted as $A = [a_1, \cdots, a_r]^T$, where $A \in \mathbb{R}^{r \times d_{in}}$. 
Following this, the CFS router employs a weight tensor $W_{AB} \in \mathbb{R}^{r \times d_{in} \times E}$ to map each factor $A[i] \in \mathbb{R}^{1 \times d_{in}}$ in $A$ to an expert dimension $E$. 
The mapping process for each factor $A[i]$, normalized via softmax and aggregated across $r$ factors, is given by:
% \begin{gather+}[1]
\begin{align}
    R^B_{\text{CFS}}(A) = \sum_{i=1}^{r} \text{softmax}(A[i] \cdot W_{AB}[i]),
\end{align}
% \end{gather+}
where $W_{AB}[i] \in \mathbb{R}^{d_{\text{in}} \times E}$ is the mapping matrix associated with $A[i]$.

The factors selection for LoRA $B$ integrates outputs from both the IFS $R^B_{\text{IFS}}(\cdot)$ and CFS $R^B_{\text{CFS}}(\cdot)$ routers via a late fusion
% ~\cite{ramachandram2017deep} 
strategy, forming the selection vector $g_B$ as follows:
% \begin{gather+}[0.8]
\begin{align}
    p^B_{\text{IFS}} = \text{softmax}(W^B_{\text{IFS}} \cdot R^B_{\text{IFS}}(h)) \\
    % p^B_{\text{CFS}} &= \text{softmax}(W^B_{\text{CFS}} \cdot R^B_{\text{CFS}}(A)) \\
    p^B_{\text{CFS}} = \text{softmax}(R^B_{\text{CFS}}(A)) \\
    g_B = \text{top}_r(p^B_{\text{IFS}} + p^B_{\text{CFS}}),
\end{align}
% \end{gather+}
where, $R^B_{\text{IFS}}(h) \in \mathbb{R}^{d_{out}}$ for LoRA $B$ is computed similarly to $R^A_{\text{IFS}}(h)$. 

The probability distribution $p^B_{\text{IFS}}  \in \mathbb{R}^E$ and $p^B_{\text{CFS}}  \in \mathbb{R}^E$ reflect the selections from the independent and conditional routers, respectively. The final selection vector $g_B \in \mathbb{R}^E$ is determined by combining these distributions and identifying the top $r$ factors.
This CFS strategy enables the selection for LoRA $B$ to be informed by factors selected for LoRA $A$, fostering a more cohesive selection process.

\subsubsection{Reconstruction of Dynamic Adaptation Matrices}

Finally, \shortmodelname{} constructs dynamic adaptation matrices by leveraging the factor selection vectors $g_A$ and $g_B$, gathering the chosen factors $\{a_k, b_k\}_{k \in K}, |K|=r$, to assemble the final matrices for LoRA $A$ and $B$. 
Consequently, in each forward pass, the weight adjustment matrix $\Delta W \in \mathbb{R}^{d_{out} \times d_{in}}$ is dynamically calculated based on these selected factors, formulated as:
% \begin{gather+}[0.8]
\begin{align}
 \Delta W = BA = [b_1,\cdots, b_r] [a_1,\cdots, a_r]^T.
\end{align}
% \end{gather+}

%% file: figures/routing.tex
\begin{figure}[!t]
    \centering
\includegraphics[width=\linewidth]{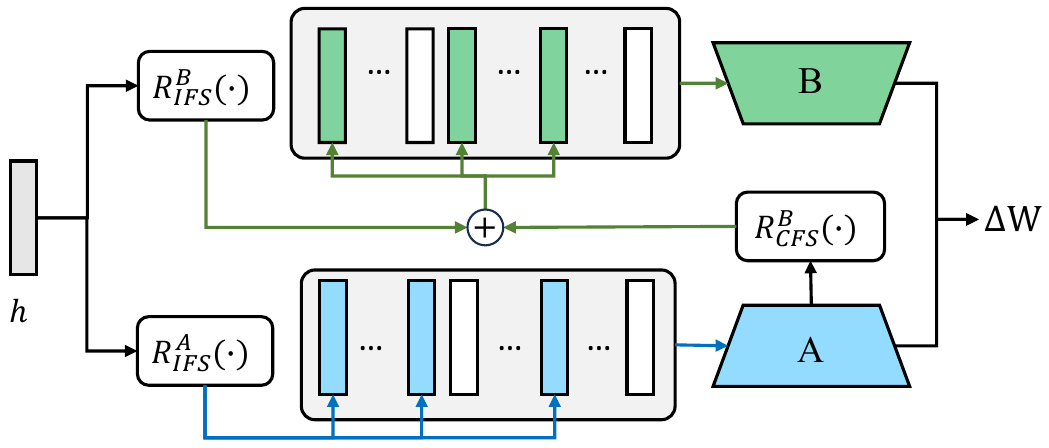}
    \caption{\textbf{Dynamic Factor Selection in \shortmodelname{}.} 
    \shortmodelname{} treats low-rank decomposition factors as experts and dynamically constructs the LoRA $A$ and $B$ through two independent routers 
    $R_{\text{IFS}}^A(\cdot)$ and $R_{\text{IFS}}^B(\cdot)$, complemented by a conditional router $R^B_{\text{CFS}}(\cdot)$.
    }
    \label{fig:routing}
\end{figure}

%% file: sec/5_exp.tex
\input{tables/main_results_full_cfs}

\section{Experimental Methodology}

\subsection{Datasets}

\paragraph{Training Datasets} 
To validate the effectiveness of \shortmodelname{}, we perform instruction tuning on \textbf{Vision-Flan}~\cite{visionFlan2023}, a human-annotated multimodal instruction tuning dataset with 187 diverse tasks. Its diversity in visual instruction tasks makes it ideal for investigating task interference. To minimize computational cost, we utilize a scaled-down version with up to 1,000 instances per task, totaling 182,167 instances.

\paragraph{Evaluation Datasets}
We evaluate our method on \textbf{MME} \cite{fu2023mme}, a comprehensive multi-modal evaluation benchmark measuring both perception and cognition abilities across 14 subtasks, to assess the overall capabilities of \shortmodelname{}. 
Alongside MME, we further probe the model's various capabilities using 7 multimodal datasets.
For Optical Character Recognition, we utilize \textbf{Text-VQA}~\cite{textvqa}, and for reasoning, we employ the Visual Spatial Reasoning (\textbf{VSR}) dataset~\cite{VSR}. 
Perception capability is tested on \textbf{CIFAR-10/100}~\cite{krizhevsky2009learning} and \textbf{MNIST}~\cite{lecun1998mnist}, following the guidance of \cite{zhai2023investigating}. The \textbf{SNLI-VE} dataset~\cite{xie2019visual} evaluates the Visual Entailment capabilities, and \textbf{POPE}~\cite{li2023evaluating} examines the tendency to objects hallucination.

\subsection{Evaluation Metrics}

For MME scores, we employ the official evaluation tool\footnote{https://github.com/BradyFU/Awesome-Multimodal-Large-Language-Models/tree/Evaluation}, aggregating the Perception and Cognition metrics. 
For other multimodal datasets, we leverage Vicuna 1.5 13B~\cite{vicuna2023}, the state-of-the-art open-source LLM to assess the accuracy of each prediction compared with ground-truth target output. More details are in Appendix \ref{appx:eval_metric}.

%% file: tables/main_results_full_cfs.tex
\begin{table*}[ht]
  \centering

  \resizebox{0.95\textwidth}{!}{%
  % \tiny
  \begin{NiceTabular}{l  | c c | c  c c c c c c c c}
  \toprule
   \textbf{Model}  & Factors & Rank & \Block[tikz={fill=gray!30}]{*-1}{} \textbf{MME} & Text-VQA & VSR & SNLI-VE & CIFAR-10 & CIFAR-100 & MNIST &
    Pope  & \Block[tikz={fill=gray!30}]{*-1}{} \textbf{MMAvg} \\
     \midrule
  LLaVA$_{\text{Align}}$ & -   & - & 1110.82 & 32.62 & 50.16 & 34.51 & 80.00 & 58.04 & 52.79 & 59.10 & 52.46\\
  LLaVA$_{\text{FT}}$ & -   & - & 1587.26 & 37.26 & 53.76 & 43.35 & 92.97 & 63.73 & 94.27 & 80.82 & 66.59 \\
   \midrule
  LoRA & -    & 2 & 1291.20 & \textbf{39.86} & 51.88 & 31.80 & 85.51 & \textbf{49.23} &  79.22 & 76.72 & 59.17 \\
  LoRA & -   & 4 & 1345.86 & 39.44 & 53.19 & 33.08 &  86.62 & 47.36 & 80.89 & \textbf{76.89} & 59.64\\
  LoRA & -  & 8 &  1312.87 & 39.20 & 53.27 & 36.36 & 88.92 & 46.88 &  82.95 & 75.48 & 60.44 \\
  LoRA  & - & 16 & 1381.23 & 39.22 & \textbf{53.60} & 36.11 & 87.31 & 45.60 &  \textbf{85.92} & 75.16 & 60.42\\
  LoRA  & -  & 32 & \textbf{1393.67} & 39.20  & 52.95 & \textbf{44.56} & \textbf{90.10} & 45.90 &  83.42 & 72.33 & \textbf{61.21}\\
   \midrule
  % \shortmodelname{}  & \xmark & 16 & 2 & 1401.37 & \second{39.86} & 51.06 & 32.77 & 89.50 & 51.34  & 77.83 & 76.13 & 59.78  \\
  \shortmodelname{}  & 16 & 2 & 1417.83 & 39.82 & 52.13 & \textbf{35.38} & 90.14 & \textbf{58.05} & \textbf{85.98} & 73.86 & 62.19 \\
  % 1414.78 & 38.74 & \second{51.96} & \textbf{40.56} & \second{91.10} & \second{53.68}  & \textbf{86.85} & \second{77.01} &  \second{62.84} \\
  % \shortmodelname{}  & \xmark & 32 & 2 & \textbf{1480.88} & 38.90 & 50.82 & 33.24 & 90.86 & 51.11  & 82.98 & 75.90 & 60.54 \\
  \shortmodelname{}  & 32 & 2 & \textbf{1459.15} & \textbf{40.46} & \textbf{52.62} & 35.04 & \textbf{91.02} & 57.95 & 85.26 & \textbf{78.31} & \textbf{62.95} \\
  % \textbf{1482.64}  & \textbf{40.88} & \textbf{54.09} & \second{35.88} & \textbf{91.47} & \textbf{58.41} & \second{85.66} & \textbf{77.06} & \textbf{63.35} \\
  \midrule
  % \shortmodelname{}  & \xmark & 16 & 4 & 1428.61 & 39.76 & 50.82 & 35.09 & \textbf{92.87} & \textbf{60.88} & 86.76 & 77.36 & 63.36 \\
  \shortmodelname{}  & 16 & 4 & 1443.82  & \textbf{40.66} & \textbf{52.70} & \textbf{43.10} & \textbf{91.59} & 57.28 & 85.25 & 78.13 & \textbf{64.10} \\
  % \second{1474.70} & 40.30 & 53.19  & \second{36.00} & \textbf{92.90} & \second{60.52} & 85.33 & 77.24 & 63.64 \\
  % \shortmodelname{}  & \xmark & 32 & 4 & 1445.62 & \textbf{41.48} & \textbf{55.07} & \second{40.49} & \second{92.06} & \second{59.69} &  \second{87.59} & 77.72 & \textbf{64.87} \\
  \shortmodelname{}  & 32 & 4 & \textbf{1509.61} & 40.42 & 49.18 & 36.69 & 91.40 & \textbf{59.27} & \textbf{87.68} & \textbf{78.48} & 63.30 \\
  % \textbf{1497.20} & \second{40.78} & \second{54.50} & 34.40 & 91.08  & 59.89 &\textbf{ 88.21} & \textbf{78.56} & \second{63.92}  \\
   \midrule
  % \shortmodelname{}  & \xmark & 16 & 8 & 1463.49 & 39.46 & 49.59 & 39.41 & \second{92.81} & \second{52.17} &  81.72 & \second{78.11} & 61.90 \\
  \shortmodelname{}  & 16 & 8 & 1485.26  & 39.92 & \textbf{52.70} & \textbf{40.74} & \textbf{92.85} & 53.96 & 82.95 & 75.31 & 62.63 \\
  % 1478.97 & \second{40.16} & 51.96 & 34.80 & 92.52 & \textbf{57.90} & 83.51 & 77.53 & \second{62.63} \\
  % \shortmodelname{}  & \xmark & 32 & 8 & \textbf{1489.16} & 39.90 & \textbf{53.52} & \second{40.38} & 91.05 & 50.53 & \second{84.47} & 77.66 & 62.50   \\
  \shortmodelname{}  & 32 & 8 & \textbf{1485.48}  & \textbf{40.02} & 51.15 & 37.77 & 91.12 & \textbf{60.25} & \textbf{86.64} & \textbf{78.87} & \textbf{63.69}\\
  % \textbf{1540.74} & \textbf{40.94} &  \textbf{53.76} & \textbf{42.25}  & \second{92.77} & 48.94  & \second{84.16} & \textbf{79.36} & \textbf{63.17} \\

  \bottomrule
  \end{NiceTabular}}
  \caption{ \textbf{Zero-shot Multi-modal Evaluation.} 
  %We report MME scores using the official evaluation tool, encompassing both Perception and Cognition metrics.
  %For other multimodal datasets, Accuracy is assessed with Vicuna-1.5. 
  LLaVA$_{\text{Align}}$ indicates the stage-one LLaVA-v1 with only feature alignment but not visual instruction tuning, and LLaVA$_{\text{FT}}$ is the fully fine-tuned LLaVA using the same Vision-Flan dataset. The \textbf{MMAvg} column denotes the average performance across seven multimodal datasets, except for MME. 
  % All tasks are unseen to the model during training, except for A-OKVQA$^*$.
  The best performance is in \first{bold}. 
  % \lifu{we also need to report the performance of LLaVA without LoRA} \ying{Updated}
  }
  \label{table:main}
  \end{table*}

%% file: sec/6_results.tex
\section{Results and Discussion}

\paragraph{Comparison with LoRA}
We first present a detailed comparison between \shortmodelname{} and the conventional LoRA, focusing on their performance in MME and 7 other multimodal tasks, as detailed in Table \ref{table:main}. We observe that \shortmodelname{} consistently surpasses LoRA when both models operate at the same ranks on both MME and the additional multimodal tasks, and even demonstrate superior performance when compared to LoRA with a higher rank number.
For instance, \shortmodelname{} (with rank $r$=2 and factors $E$=16) outperforms LoRA (rank $r$=32) by 1.7\% in MME and 1.6\% on average across other multimodal evaluations.

\paragraph{Increase the Number of Rank}
We investigate the impact of increasing the rank number while keeping the number of factors constant. As shown in Table \ref{table:main}, \shortmodelname{} exhibited a notable performance enhancement as the rank number increased from 2 to 4, when the factor number was fixed.
Specifically, increasing the rank $r$ from 2 to 4 leads to a performance uplift of 1.8\% in MME and 3.1\% in MMAvg with $E=16$ factors, and a 3.5\% improvement in MME and a 0.6\% increase in MMAvg with $E=32$ factors. 
However, further increasing the rank to 8 shows diminishing returns in performance gains. We hypothesize this decline might potentially be due to the expanded combination pool for constructing the adaptation matrices.

\paragraph{Increasing the Number of Factors} 
In scenarios where the rank number is held constant, our findings reveal a general trend of performance improvement for \shortmodelname{}, as shown in Table \ref{table:main}. 
This improvement can be attributed to the model's increased capacity for providing a richer set of factors to tailor the model to specific multimodal tasks.

\paragraph{The Effect of Routing Strageties}
\input{tables/routing_strategy}

In this experiment, we examine different routing strategies for the IFS router. In particular, we implement the Task-Specific Routing paradigm which leverages the definition of each multimodal instruction task to inform the selection of decomposition factors (details can be found in Appendix \ref{appx:task_router}).
Table \ref{table:routing_strategy} shows that Instance-based Routing significantly outperforms Task-specific routing, achieving a higher MME score and average performance across the additional multimodal tasks. The superior performance of Instance-based Routing likely stems from its inherent flexibility. Unlike Task-specific Routing, which has the same selection of factors at different layers for inputs from the same task, Instance-based Routing adapts its selection based on the varying hidden states from previous layers, leading to a more flexible routing mechanism.

Furthermore, we investigate whether the superior performance is due to the introduction of extra expert parameters and not the routing mechanism. Table \ref{table:routing_strategy} reports the comparison with a random routing baseline, which randomly selects $r$ factors. Our observations reveal that both Instance-based Routing and Task-specific routing surpass the random baseline, suggesting that the routing mechanism, rather than the inclusion of additional expert parameters, is responsible for the performance enhancements.

\input{figures/cfs_score_comp}
\paragraph{Impact of Conditional Factor Selection}

We assess the impact of Conditional Factor Selection (CFS) through an ablation analysis, comparing \shortmodelname{}'s averaged performance with and without the CFS across seven multimodal datasets. The comparative results, as shown in Figure~\ref{fig:cfs_comp} demonstrate that incorporating the CFS router in general consistently improves the performance across different factor and rank settings. This enhancement is hypothesized to stem from the CFS's role in strengthening the interdependency between the factor selections of LoRA $A$ and $B$.

\paragraph{Factor Selection Pattern on Unseen Tasks}

Our analysis delves into the factor selection patterns of LoRA $A$ for unseen multimodal tasks.
We randomly sample 300 instances from each of seven unseen multimodal tasks and visualize the factor selection within the MLP layer using t-SNE \cite{van2008visualizing}, as shown in Figure \ref{fig:t_sne}. 
We observe that instances from identical tasks tend to cluster, indicating the effectiveness of an instance-based routing strategy in assigning diverse factor sets across tasks.

Furthermore, we visualize the factor selection patterns for similar seen and unseen tasks. We pair five distinct unseen tasks, each probing a different capability, with five similar seen tasks from the training set: SNLI-VE (unseen) with Image-Text (seen) for assessing visual entailment, Text-VQA (unseen) with InfoGraphicVQA (seen) for OCR capabilities, VSR (unseen) with GQA (seen) for reasoning, Pope (unseen) with VQA-Object-Presence (seen) for hallucination detection, and CIFAR-10 (unseen) with ExDark (seen) for perception capabilities.
The t-SNE visualization shown in Figure \ref{fig:seen_unseen_factor} depicts the distribution of factor selection across MLP layers, with the first row in the legend indicating the seen tasks, and the second row denoting the corresponding unseen tasks. Similar color schemes are used for each pair of similar seen and unseen tasks for clarity.
Our observations reveal that \shortmodelname{} effectively activates factors analogous to those employed in similar training tasks. This finding suggests that the model can adapt its factor selection strategies to new, unseen tasks based on its training on similar seen tasks.

\input{figures/factor_selection_fig}
\input{figures/seen_unseen_factor}

\paragraph{Analysis of Task Interference}

\input{tables/training_tasks}

\input{figures/task_interference_comp.tex}

To assess \shortmodelname{}'s efficacy in mitigating task interference, we test it on the same six training tasks: 
 ``ScienceQA'', ``COCO'', ``FairFace'', ``iNaturalist'', ``ST-VQA'', and ``PACS'', discussed in Section \ref{subsec:task_inteference}.
 For each task, we randomly sample 300 instances not included in the instruction-tuning phase for evaluation. We compare \shortmodelname{} against both the conventional LoRA and task-specialized LoRA models (LoRA$_\text{Specialist}$) that are fine-tuned with task-specific adaptation parameters for each task. Table \ref{table:training} shows that conventional LoRA models exhibit varying degrees of performance degradation across tasks when compared to LoRA$_\text{Specialist}$. In contrast, \shortmodelname{} suffers less from performance degradation and demonstrates more consistent and robust performance across different tasks, suggesting its effectiveness in reducing task interference.

Moreover, we visualize the task interference scores using Equation \ref{eq:L} and \ref{eq:I}. Given that \shortmodelname{} dynamically selects a subset of factors ($r$ out of $E$) for different instances, we record gradients concerning all $E$ factors and compare the task interference scores between standard LoRA models (with $r$=16) and \shortmodelname{} (with $E$ = 16 and $r$ = 4). Figure \ref{fig:ti_comp} visualizes the interference scores for both LoRA $A$ and LoRA $B$ aggregated across all adaptation layers, including MLP and self-attention layers.  
The analysis reveals that \shortmodelname{} ($E$=16, $r$=4) exhibits lower negative interference scores compared to the standard LoRA ($r$=16), underscoring \shortmodelname{}'s efficacy in reducing task interference.

%% file: tables/routing_strategy.tex
% \begin{table}[!t]
%   \centering
%   \resizebox{\linewidth}{!}{%
%   % \tiny
%   \begin{tabular}{l c c c | c c}
%   \toprule
%    \textbf{Model} & Routing  & Factors & Rank & \textbf{MME}  & \textbf{MMAvg} \\
%    \midrule
%    \shortmodelname{} & Random  & 32 & 4 & 1007.40 & 49.12  \\
%   \midrule
%   \shortmodelname{} & Instance  & 32 & 4 & \textbf{1445.62} & \textbf{64.87} \\
%   \shortmodelname{} & Task  & 32 & 4 & 1430.31 & 62.08 \\
%   % 39.04 & 50.74 & 33.92 & 91.34 & 61.48 & 77.91 & 78.03 & 80.06 & 64.06

%   \bottomrule
%   \end{tabular}
%   }
%   \caption{ \textbf{Comparison between Various Routing Strategies.} The \textbf{MMAvg} column denotes the average performance across seven multimodal datasets.}
%   \label{table:routing_strategy}
%   \end{table}

\begin{table}[!t]
  \centering
  \resizebox{\linewidth}{!}{%
  % \tiny
  \begin{tabular}{l c c c | c c}
  \toprule
   \textbf{Model} & Routing  & Factors & Rank & \textbf{MME}  & \textbf{MMAvg} \\
   \midrule
   \shortmodelname{} & Random  & 32 & 4 & 1007.40 & 49.12  \\
  \midrule
  \shortmodelname{} & Instance  & 32 & 4 & \textbf{1509.61} & \textbf{63.30} \\
  \shortmodelname{} & Task  & 32 & 4 & 1381.87 & 61.75 \\

  \bottomrule
  \end{tabular}
  }
  \caption{ \textbf{Comparison between Various Routing Strategies.} The \textbf{MMAvg} column denotes the average performance across seven multimodal datasets.}
  \label{table:routing_strategy}
  % \vspace{-4mm}
  \end{table}

%% file: figures/cfs_score_comp.tex
\begin{figure}[!t]
    \centering
\includegraphics[width=\linewidth]{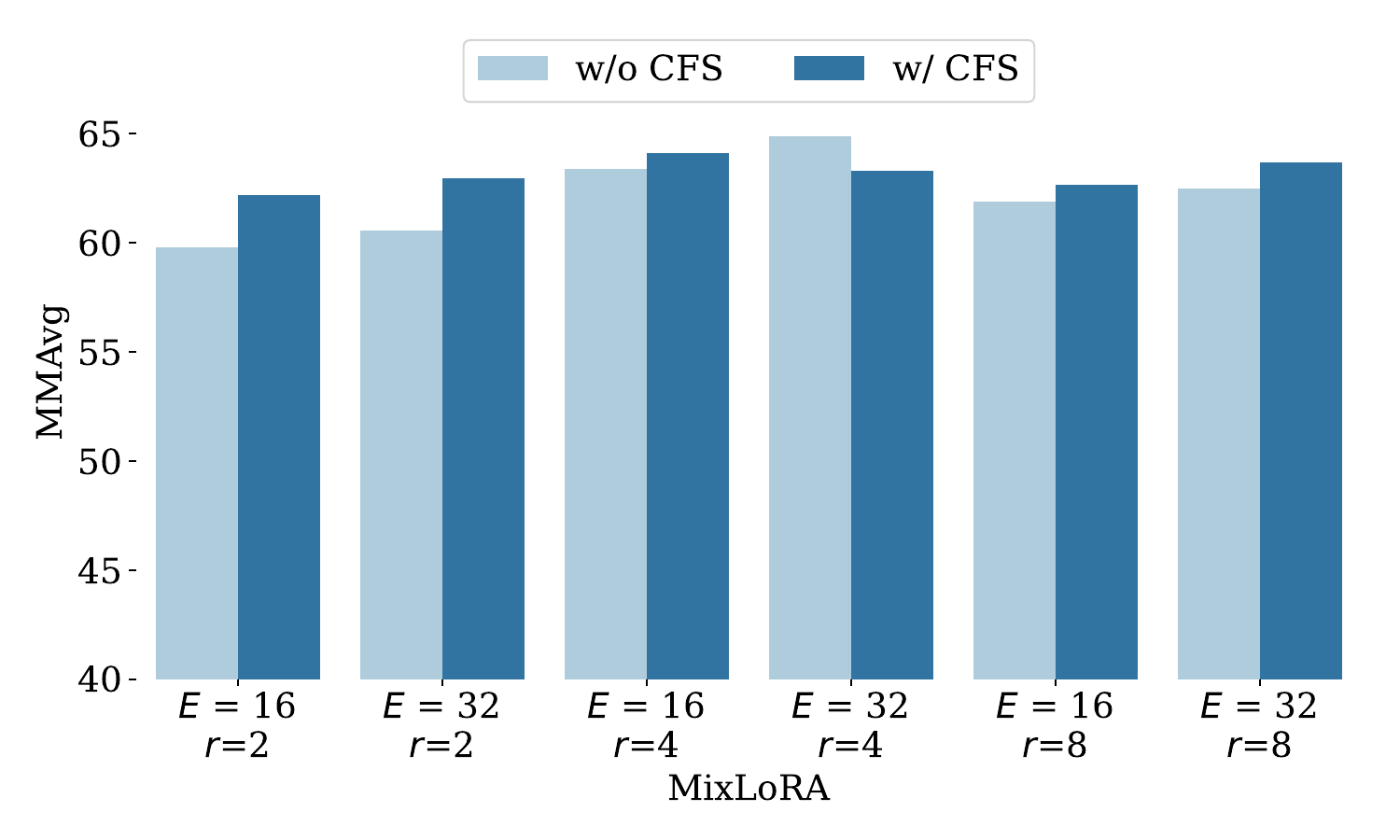}
    \caption{\textbf{Effect of Conditional Factor Selection}} 
    \label{fig:cfs_comp}
\end{figure}

%% file: figures/factor_selection_fig.tex
\begin{figure}[!t]
    \centering
        \includegraphics[width=\linewidth]{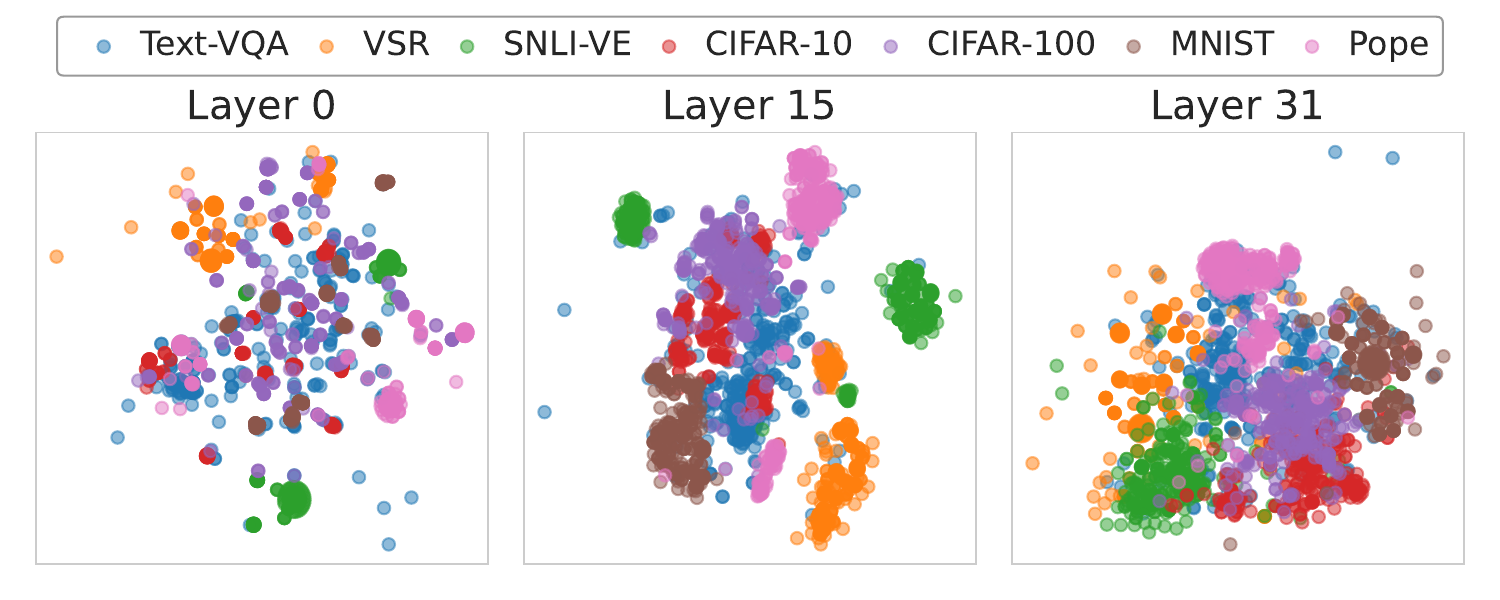}
    \caption{\textbf{T-SNE Visualization of Factor Selection Distribution for \shortmodelname{} ($E$ = 32, $r$ = 8)}. Instances are represented as points, where instances from the same task share a common color.}
    \label{fig:t_sne}
\end{figure}

%% file: figures/seen_unseen_factor.tex
\begin{figure}[!t]
    \centering
        \includegraphics[width=\linewidth]{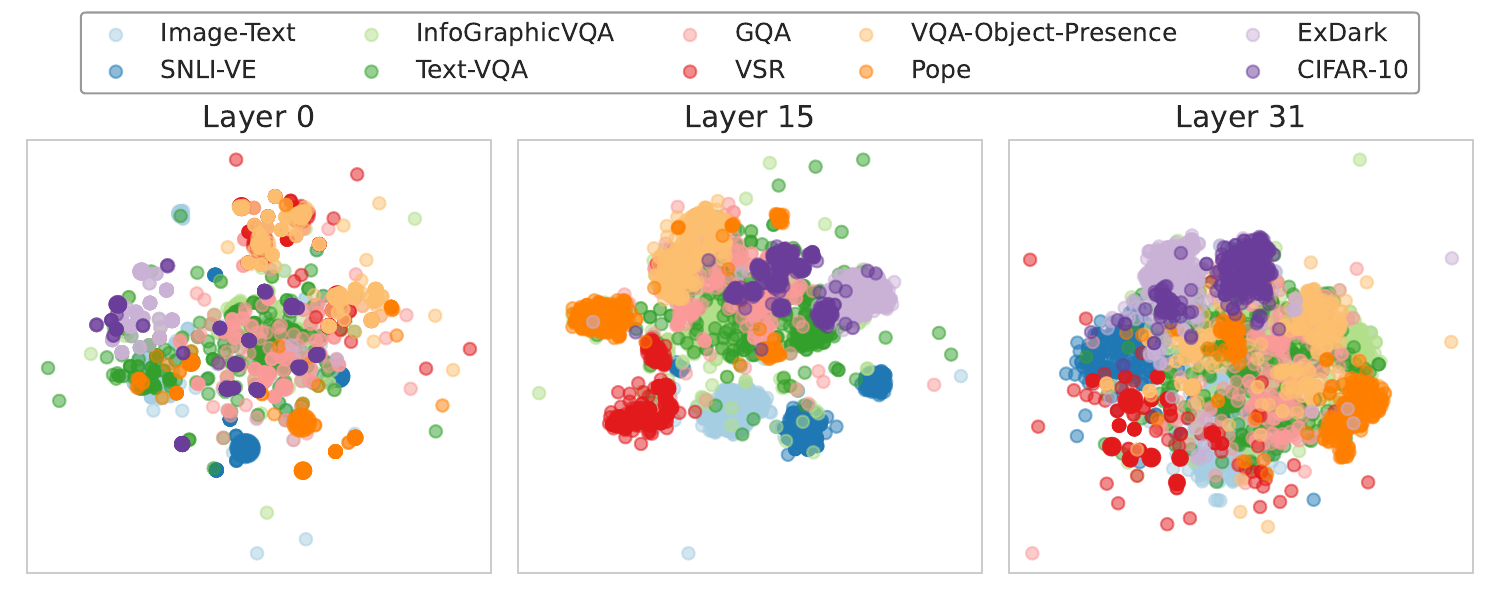}
    \caption{\textbf{T-SNE Visualization of Factor Selection in \shortmodelname{} ($E$ = 32, $r$ = 8) for Seen and Unseen Tasks}. 
    % Instances are represented as points, where instances from the same task share a common color. 
    Seen tasks (Image-Text, InfoGraphicVQA, GQA, VQA-Object-Presence, CIFAR-10) in the first row are color-matched with their unseen counterparts (SNLI-VE, Text-VQA, VSR, Pope, ExDark) in the second row.}
    \label{fig:seen_unseen_factor}
\end{figure}

%% file: tables/training_tasks.tex
\begin{table*}[!t]
  \centering
  \resizebox{0.9\linewidth}{!}{%
  % \tiny
  \begin{tabular}{l c c | c c c c c c c c  }
  \toprule
   \textbf{Model} & Factors & Rank & ScienceQA & COCO & FairFace & iNaturalist & ST-VQA &  PACS & \textbf{AVG} \\

     \midrule

  LoRA$_\text{Specialist}$  & - & 4 & 64.33	&77.67	&54.67	&58.67	&44.67&	99.00&	66.50 \\
  LoRA$_\text{Specialist}$  & - & 16 & 67.33&	76.33&	59.00	&60.00&	46.33	&99.00&	68.00\\
   \midrule
  LoRA & - & 4 & 57.67 & 76.33& 	59.67& 	57.00 & 42.33& 	99.33& 65.39 \\
  LoRA & - & 16 & 59.67 & 73.00 & 59.33 & 58.33 & 43.67 & 99.00 & 65.50 \\
  \midrule
 \shortmodelname{} & 16 & 4 & 60.67 & 78.67 & 59.00 & 61.00 & 44.33 & 99.33 & 67.17 \\

  \bottomrule
  \end{tabular}}
  \caption{ \textbf{Multi-modal Evaluation on Seen Tasks.} 
LoRA$_\text{Specialist}$ represents the specialist LoRA model fine-tuned for each seen task individually. The \textbf{AVG} column denotes the average performance across six seen tasks.
  }
  \label{table:training}
  \end{table*}

%% file: figures/task_interference_comp.tex
\begin{figure}[t]
    \centering
    \includegraphics[width=\linewidth]{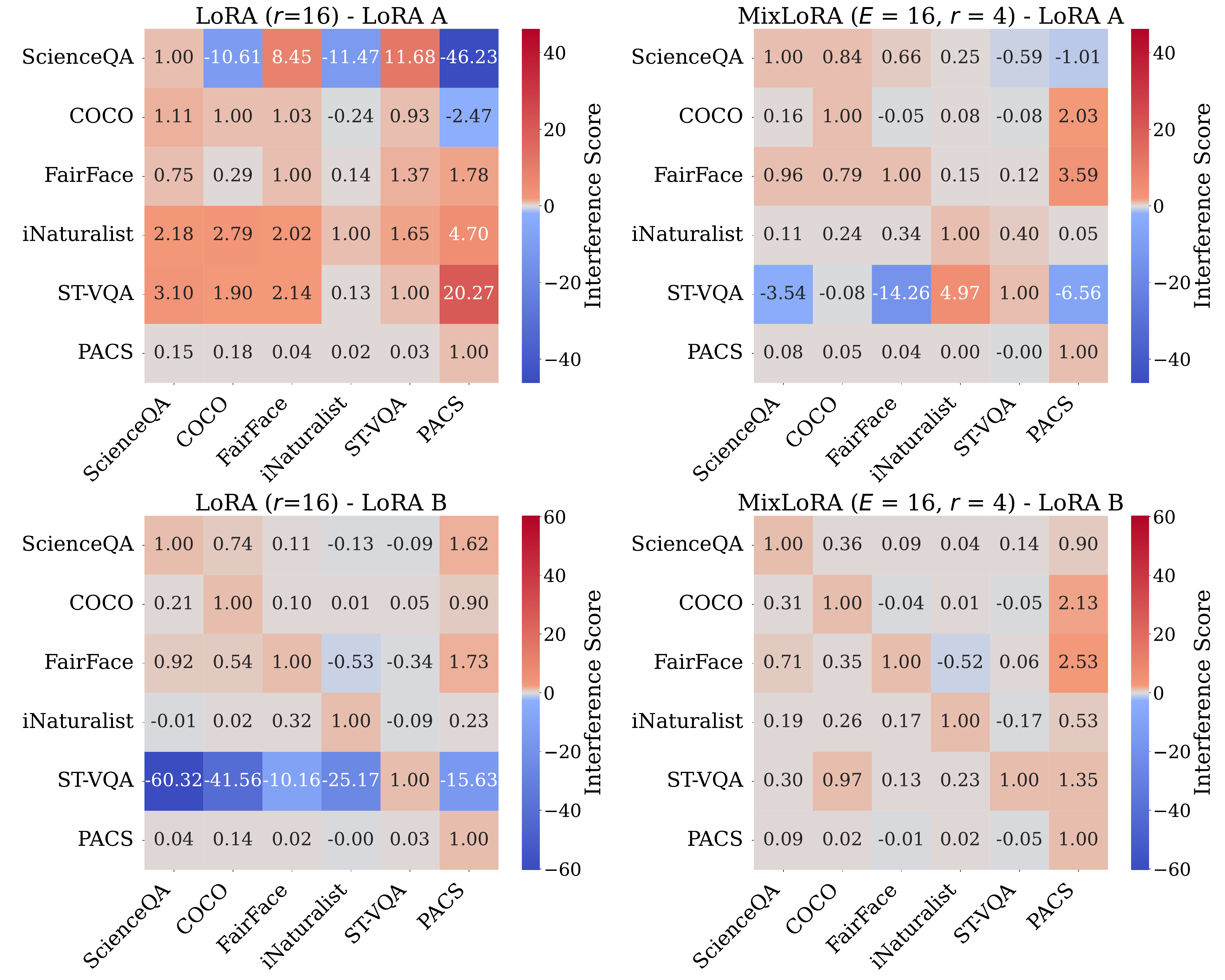}
    \caption{
    \textbf{
    The Comparison of Task Interference Score $\mathcal{I}$ between LoRA ($r$=16) and MixLoRA ($E$ = 16, $r$ = 4).} Each cell in the heatmap corresponds to the average interference score $\mathcal{I}_{i,j}$ of task $j$ (column) on the task $i$ (row) averaged across all adaption layers.
    }
    \label{fig:ti_comp}
\end{figure}

%% file: sec/7_conclusion.tex
\section{Conclusion}

We introduce \modelname{}, an innovative strategy that dynamically constructs low-rank adaptation matrices specific to individual inputs, to mitigate task interference during parameter-efficient multimodal instruction tuning. 
Comprehensive experiments across a variety of multimodal datasets have demonstrated the efficacy of \shortmodelname{}, showcasing an enhanced performance on unseen multimodal tasks compared to conventional LoRA and demonstrating its effectiveness in mitigating task interference.

%% file: sec/8_limitations.tex
\section{Limitations}

 Our study focuses on task interference within parameter-efficient multimodal instruction tuning, specifically for image and text modalities, leaving the integration of other modalities like sound and 3D point clouds as an avenue for future work. Moreover, due to the cost of training large models, our experimentation was conducted on a scaled-down version of Vision-Flan. Future studies could benefit from evaluating the effectiveness of \shortmodelname{} when applied to more extensive multimodal instruction-tuning datasets. Additionally, our method introduces extra training overhead compared to standard LoRA of the same rank.

%% file: sec/9_acknowledgements.tex
\section*{Acknowledgments}

This research is based upon work partially supported by the U.S. DARPA ECOLE Program \#HR001122S0052, FoundSci Program \#HR00112490370, and the Amazon - Virginia Tech Initiative for Efficient and Robust Machine Learning. The views and conclusions contained herein are those of the authors and should not be interpreted as necessarily representing the official policies, either expressed or implied, of DARPA or the U.S. Government. The U.S. Government is authorized to reproduce and distribute reprints for governmental purposes notwithstanding any copyright annotation therein.

%% file: sec/X_suppl.tex
\clearpage

\section{Task-Specific Routing}
\label{appx:task_router}

The Task-Specific Routing paradigm leverages the distinct characteristics of each multimodal instruction task to inform the selection of decomposition factors. 
This strategy utilizes the detailed task definition, which includes a comprehensive description of the task's requirements and the specific skills or modalities needed to successfully perform the task. 
For instance, consider the task ``OK-VQA''~\cite{marino2019ok}, the task definition is: ``\textit{Answer the question in natural language based on the content of the image. The questions require external knowledge to answer}.'' The task-specific routing strategy is formulated as:
\begin{align}
    R^A_{\text{IFS}}(z) = \text{Avg}(f_{\phi}(z)),
\end{align}
where $f_{\phi}(\cdot)$ denotes a pre-trained Large Language Model (LLM) parameterized by $\phi$, responsible for encoding the task definition $z$. 

\section{Implementation Details}

We leverage the stage-one LLaVA-v1~\footnote{https://github.com/haotian-liu/LLaVA/} (before the visual instruction tuning stage) as our pre-trained large multimodal models, specifically employing LLaVA with Vicunna-7B v1.3. For all model variants, we fine-tune this stage-one LLaVa on the scale-down version of Vision-Flan for three epochs, using a total batch size of 128 and a learning rate of $4e-5$. The fine-tuning process for \shortmodelname{} ($E$=16, $r$=4) takes approximately 20 hours on 4 A100 GPUs, with an effective batch size of 8 per GPU and a gradient accumulation step of 4.
For LoRA, we set the hyper-paramter $\alpha$ in Equation \ref{eq:lora} to be 2 $\times$ rank $r$ and for \shortmodelname{}, we define $\alpha$ as 2 $\times$ factors $|E|$. 
For the other configuration, we adopt LLaVA's default setting for LoRA fine-tuning, as provided in its codebase. For the task-specific routing, we adopt the Vicunna \cite{vicuna2023} as our pre-trained large language model $f_{\phi}(\cdot)$ for encoding task definition. Notably, Vicuna also serves as the language backbone of the LLaVA model. Following a similar approach to LoRA, for the LLaVA model with 32 Transformer layers, we insert \shortmodelname{} into all linear layers within the Transformer layers. During training, all parameters in the \shortmodelname{} module are updated, while the rest of LLaVA's parameters remain frozen.

\section{Evaluation Metrics}
\label{appx:eval_metric}

To evaluate the model performance on unseen multimodal datasets, we leverage Vicuna 1.5 13B~\cite{vicuna2023}, the state-of-the-art open-source LLM to perform the evaluation. 
Specifically, we craft a prompt template that directs Vicuna to assess the accuracy of each prediction, considering the given task instructions and ground-truth target output. 
The prompt template used is as follows: 
``
\textit{A chat between a curious user and an artificial intelligence assistant. The assistant gives helpful, detailed, and polite answers to the user's questions. 
USER: Decide if the prediction is correct given the question and the answer.
Questions: \{Question\}
Answer: \{Ground-truth Answer\}
Prediction: \{Prediction\}
Your response should only be Yes or No. ASSISTANT:}''
In this template, placeholders such as ``\{Question\}'', ``\{Ground-truth Answer\}'', and ``\{Prediction\}'' will be substituted with the specific details of each test instance. 
If Vicuna determines the prediction is correct, it outputs ``Yes'', and ``No'' otherwise. As all tasks are classification tasks, we compute accuracy based on Vicuna's judgments. 